%% file: warp.tex
\documentclass[english]{article}
\usepackage{times}
\usepackage{amsmath}
\usepackage{amsthm}
\usepackage{amssymb}
\usepackage{amscd}
\usepackage{natbib}
\usepackage{hyperref}
\usepackage{cleveref}
\usepackage[accepted]{icml2017}

\usepackage{esint}
\usepackage{babel}
\usepackage{booktabs}
\usepackage{todonotes}
\usepackage{algorithm}

\usepackage{pgf,tikz}
\usetikzlibrary{matrix,backgrounds,fit}
\usetikzlibrary{arrows}
\usepackage{pgfplots}

\newtheorem{lemma}{Lemma}

\newcommand{\R}{\mathbb{R}}
\newcommand{\Z}{\mathbb{Z}}

\newcommand{\conv}[1]{\underset{#1}{\ast}}

\newcommand{\ItoG}[1]{\tilde{#1}}
\newcommand{\wideItoG}[1]{\widetilde{#1}}
\newcommand{\Ig}{\ItoG{I}}
\newcommand{\Fg}{\ItoG{F}}

\newcommand{\Iw}{I_\textrm{w}}
\newcommand{\Fw}{F_\textrm{w}}

\renewcommand{\paragraph}[1]{\medskip\par\noindent\textbf{#1}}

\renewcommand\[{\begin{equation}}
\renewcommand\]{\end{equation}}
\renewcommand{\cite}{\citep}

\begin{document}
\twocolumn[
\icmltitle{Warped Convolutions: Efficient Invariance to Spatial Transformations}

\begin{icmlauthorlist}
\icmlauthor{Jo\~{a}o F.~Henriques}{ox}
\icmlauthor{Andrea Vedaldi}{ox}
\end{icmlauthorlist}
\icmlaffiliation{ox}{Visual Geometry Group, University of Oxford, United Kingdom}
\icmlcorrespondingauthor{Jo\~{a}o F.~Henriques}{joao@robots.ox.ac.uk}

\vskip 0.3in
]

\printAffiliationsAndNotice{}

\begin{abstract}
Convolutional Neural Networks (CNNs) are extremely efficient, since they exploit the inherent translation-invariance of natural images. However, translation is just one of a myriad of useful spatial transformations. Can the same efficiency be attained when considering other spatial invariances?
Such generalized convolutions have been considered in the past, but at a high computational cost.
We present a construction that is simple and exact, yet has the same computational complexity that standard convolutions enjoy. It consists of a constant image warp followed by a simple convolution, which are standard blocks in deep learning toolboxes.
With a carefully crafted warp, the resulting architecture can be made equivariant to a wide range of two-parameter spatial transformations. We show encouraging results in realistic scenarios, including the estimation of vehicle poses in the Google Earth dataset (rotation and scale), and face poses in Annotated Facial Landmarks in the Wild (3D rotations under perspective).
\end{abstract}

\section{Introduction}

A crucial aspect of current deep learning architectures is the encoding of invariances. This fact is epitomized in the success of convolutional neural networks (CNN), where \emph{equivariance to image translation} is key: translating the input results in a translated output. When invariances are present in the data, encoding them explicitly in an architecture provides an important source of regularization, which reduces the amount of training data required for learning.

Invariances may also be used to improve the efficiency of implementations. For instance, a convolutional layer requires orders of magnitude less memory (by reusing filters across space) and less computation (due to their limited support) compared to a fully-connected layer. Its local and predictable memory access pattern also makes better use of modern hardware's caching mechanisms.

The success of CNNs indicates that translation invariance is an important property of images. However, this does not \emph{explain why} translation equivariant operators work well for image understanding. The common interpretation is that such operators are matched to the statistics of natural images, which are well known to be translation invariant~\cite{hyvarinen2009natural}. However, natural image statistics are also (largely) invariant to other transformations such as isotropic scaling and rotation, which suggests that alternative neural network designs may also work well with images. Furthermore, in specific applications, invariances other than translation may be more appropriate.

Therefore, it is natural to consider generalizing convolutional architectures to other image transformations, and this has been the subject of extensive study~\cite{kanazawa2014locally,bruna2013learning,cohen2016}. Unfortunately these approaches do not possess the same memory and speed benefits that CNNs enjoy. The reason is that, ultimately, they have to transform (warp) an image or filter several times~\cite{kanazawa2014locally,marcos2016learning,dieleman2015rotation}, incurring a high computational burden. Another approach is to consider a basis of filters (analogous to eigen-images) encoding the desired invariance~\cite{taco_cohen_learning_2014,bruna2013learning,cohen2016}. Although they are able to handle transformations with many pose parameters, in practice most recent proposals are limited to very coarsely discretized transformations, such as horizontal/vertical flips and 90$^{\circ}$ rotations~\cite{dieleman2015rotation,taco_cohen_learning_2014}.

In this work we consider generalizations of CNNs that overcome these disadvantages. Well known constructions in group theory enable the extension of convolution to general transformation groups~\cite{folland1995course}. However, this generality usually comes at an increased computational cost or complexity. Here we show that, by making appropriate assumptions, we can design convolution operators that are equivariant to a large class of two-parameter transformations while reducing to a standard convolution in a warped image space. The fixed image warp can be implemented using bilinear resampling, a simple and fast operation that has been popularized by spatial transformer networks~\cite{jaderberg2015spatial,heckbert1989fundamentals} and is part of most deep learning toolboxes. Unlike previous proposals, the proposed \emph{warped convolutions} can handle continuous transformations, such as fine rotation and scaling.


This makes generalized convolution easily implementable in neural networks, reusing fast convolution algorithms on GPU hardware, such as Winograd~\cite{lavin2015fast} or the Fast Fourier Transform~\cite{lyons2010understanding}.

\section{Generalizing convolution}\label{s:background}

\subsection{Discrete and continuous convolution}

We start by looking at the basic building block of CNNs, i.e.\ the convolution operator. This operator computes the inner product of an image $I\in\R^{N\times N}$ with a translated version of the filter  $F\in\R^{M\times M}$, producing a new image as output:
\begin{equation}\label{eq:discrete-conv}
(I\ast F)(j)=\sum_{k}I(k)F(j-k),
\end{equation}
where $k,\, j\in\Z^{2}$ are two-dimensional vectors of indexes, and the summation ranges inside the extents of both arrays. Over the next sections it will be more convenient to translate the image $I$ instead of the filter $F$. This alternative form of \cref{eq:discrete-conv} is obtained by the substitution $k \leftarrow j + k$:
\begin{equation}\label{eq:discrete-conv1}
(I\ast F)(j)
=\sum_{k}I(j+k)F(-k)
\end{equation}

In the neural network literature, this is often written using the \emph{cross-correlation convention}~\cite{goodfellow2016deep}, by considering the reflected filter $F_-(j)=F(-j)$:
\begin{equation}\label{eq:discrete-corr}
(I\ast F)(j)
=\sum_{k}I(j+k)F_-(k).
\end{equation}
%

To handle continuous deformations of the input, it is more natural to express \cref{eq:discrete-conv1} as an integral over continuous rather than discrete inputs:
\begin{equation}\label{eq:continuous-conv}
(I \ast F)(y)
=\int I(y+x)\,F(-x)\,dx,
\end{equation}
where $I,\, F:\,\mathbb{R}^2\rightarrow\R$ are functions of continuous inputs in $\R^{2}$. The real-valued 2D vectors $x,y \in \mathbb{R}^2$ now play the role of the indexes $k\in\Z^{2}$. \Cref{eq:continuous-conv} reduces to the discrete case of \cref{eq:discrete-conv} if we define $I$ and $F$ as the sum of delta functions on grids (Dirac comb). Intermediate values can be obtained by interpolation, such as bilinear (which amounts to convolution of the delta functions with a triangle filter \cite{jaderberg2015spatial}). Importantly, such continuous images can be deformed by a rich set of continuous transformations of the input coordinates, whereas strictly discrete operations would be more limiting.


\subsection{Convolution on groups}\label{sub:generalized-conv}

The standard convolution operator of~\cref{eq:continuous-conv} can be interpreted as applying the filter to translated versions of the image. We wish to replace translations with other image transformations $g$, belonging to a group $G$. In the context of machine learning models for images, this generalized (group) convolution can be understood to exhaustively search for a pattern at various poses $g \in G$ (e.g. rotation angles or scale factors)~\cite{dieleman2015rotation,kanazawa2014locally}.

Following standard derivations~\cite{folland1995course}, the most common way of generalising convolution to transformations other than translations starts from the Haar integral. Given a measure $\mu$ over a group $G$, one can define the integral of a real function $\Ig:G \rightarrow \mathbb{R}$, written:
$$
\int_G \Ig(g)\,d\mu(g).
$$
The (left) Haar measure is the most natural choice for $\mu$; it is the only measure (up to scaling factors) that is (left) invariant to group translation. In other words, $\mu$ satisfies the equation:
$$
\forall h \in G :\quad \int_G \Ig(hg)\,d\mu(g) = \int_G \Ig(g)\,d\mu(g).
$$
Mirroring~\cref{eq:continuous-conv}, the \emph{group convolution} of two functions $\Ig$ and $\Fg$ is defined as%
\footnote{%
Due to the Haar invariance property, this definition is equivalent to the following one, also commonly found in the literature:
$$
(\Ig \conv{G} \Fg)(t) = \int \Ig(g) \Fg(g^{-1}t)\,d\mu(g).
$$
The equivalence mirrors the one between \cref{eq:discrete-conv} and \cref{eq:discrete-conv1}, and can be easily proved:
\begin{align*}
(\Ig \conv{G} \Fg) (t) 
&= \int \Ig(tg) \Fg((tg)^{-1} t)\,d\mu(g)\\
&= \int \Ig(tg) \Fg(g^{-1})\,d\mu(g).
\end{align*}
}
\begin{equation}\label{eq:group-conv}
(\Ig \conv{G} \Fg)(t) = \int \Ig(tg) \Fg(g^{-1})\,d\mu(g).
\end{equation}
From the viewpoint of statistical learning, a key property of convolution is \emph{equivariance}. Consider the (left) translation operator
\begin{equation}
L_h(\Ig) : \quad t \longmapsto \Ig(h^{-1}t).
\end{equation}
Then:
\begin{lemma}\cite{folland1995course}\label{lemma}
Convolution is equivariant with group translations, in the sense that $L_h$ commutes with $\conv{G}$:
$$
L_h(\Ig \conv{G} \Fg) (t) = (L_h(\Ig) \conv{G} \Fg)(t).
$$
\end{lemma}
\begin{proof}
$
L_h(\Ig \conv{G} \Fg) (t) 
= \int \Ig((h^{-1}t)g) \Fg(g^{-1})\,d\mu(g)
= \int \Ig(h^{-1}(tg)) \Fg(g^{-1})\,d\mu(g)
= (L_h(\Ig) \conv{G} \Fg)(t).
$
\end{proof}

\subsection{From groups to images}\label{sub:groups-to-images}
The functions $\Ig$ and $\Fg$ above have been defined on groups. In applications, however, we are interested in \emph{images}, i.e.\ functions  $I : \Omega \rightarrow \mathbb{R}$ defined on a subset $\Omega$ of the real plane $\mathbb{R}^2$.

The connection between the two function types is easy to establish. The assumption is that $G$ acts\footnote{This means that $g$ defines a mapping $\Omega \mapsto \Omega$ and that the group multiplication $hg$ corresponds to function composition $(hg)x = h(gx)$.} on the image domain $\Omega$ (i.e.\ $G$ is a group of transformations of $\Omega$). We can then define the $\ItoG{\cdot}$ operator as:
$$
\Ig(g) = I(gx_0)
$$
where $x_0 \in \Omega$ is an arbitrary pivot point. Note that the values of $\tilde I$ depend only on the values of $I$ on the orbit of $x_0$, i.e.\ the set $Gx_0 = \{ gx_0 : g \in G \}$. Therefore we typically set the domain $\Omega$ of $I$ to be equal to $Gx_0$.

By this definition, left translation of $\Ig$ by $h$ corresponds to warping the image $I$ by the transformation $h$:
$$
L_h(\Ig)(t)
= I((h^{-1}t)x_0)
= I(h^{-1}(tx_0))  = \wideItoG{(I \circ h^{-1})}(t).
$$
We can then update \cref{eq:group-conv} to express group convolution as a function of images on the real plane:
\begin{equation}\label{eq:image-space-group-conv}
(I\conv{G}F)(t)=\int I(tgx_{0})F(g^{-1}x_{0})\,d\mu(g).
\end{equation}

\subsection{Standard convolutions with exponential maps}\label{sub:calculations}
In the definitions of~\cref{sub:generalized-conv,sub:groups-to-images}, while we could reduce the functions $\Ig$ and $\Fg$ to images $I$ and $F$, the result of convolution is still a function defined over a group. One needs therefore to understand how to represent such functions and calculate the corresponding integrals.

We note here that the simplest case is when $G$ is a Lie group parameterised by the exponential map $\exp: V \rightarrow G$, where $V$ is a subset of $\mathbb{R}^P$, in such a way that $\exp$ is smooth, bijective and additive ($\exp(u+v)=\exp(u)\exp(v)$). Then:
$$
\int_G \Ig(g)\,d\mu(g)
=
\int_V \Ig(\exp(u))\,du.
$$
Hence we can define the warped image
$$
\Iw(u) = \Ig(\exp(u)) = I(\exp(u)x_0), \quad u \in V.
$$
and group convolution reduces to the standard notion of convolution $\conv{V}$ on $V$:
\begin{align}\label{eq:warped-conv}
(\Ig \conv{G} \Fg)(\exp(v))
&=
\int_V
\Iw(v+u) \Fw(-u)\,du.
\end{align}

We refer to this standard convolution in warped space (\cref{eq:warped-conv}) as \emph{warped convolution}.

\paragraph{Discussion.} Note that the result is an image whose dimensionality $P$ is that of the vector space $V$; in the following, we mainly work in the case $P=2$. By far, the strongest requirement is that the map is additive: this is the same as requiring the transformation group $G$ to be Abelian, in the sense that transformations commute ($hg=gh$). In \cref{sec:examples} we will show a variety of useful image transformations that respect this property. The advantage of introducing this restriction is that calculations simplify tremendously, ultimately enabling a simple and efficient implementation of the operator as discussed below.

\section{Warped convolutions}\label{sec:warped-conv}

Our main contribution is to note that certain group convolutions can be implemented efficiently by a \emph{standard} convolution, by pre-warping the input image and filter appropriately. The warp is the same for any image, depending solely on the nature of the relevant transformations, and can be written in closed form. This result allows one to implement very efficient group convolutions using simple computational blocks, as shown in~\cref{sub:practical}.

\subsection{Warped convolution layer}

We can now reinterpret these results in terms of a new neural network convolutional layer. The input of the layer is an image $I$ and the learnable parameters are the coefficients of the filter $\Fw$. The output is a new ``image'' $C(I;\Fw)(v)$ defined on the vector space $V$. This image is obtained by first warping $I$ using the exponential map and then by convolving the result with $\Fw$ in the standard sense:
\begin{equation}\label{eq:generalized-conv}
C(I;\Fw)(v) = (I(\exp(\cdot)x_0) \conv{V} \Fw)(v).
\end{equation}
The most important property of this layer is equivariance: if we warp the image by the transformation $h=\exp(u)$, then the convolution result translates by $u$:
$$
C(I\circ h^{-1}; \Fw) = C(I;\Fw)(v-u),\quad h=\exp(u).
$$
Note that the output action equivalent to warping the input is simply to translate the result, as for standard convolution. The second most important property is that this operator can be implemented efficiently as the combination of warping and standard convolution.

\subsection{Implementation and intuition}\label{sub:practical}

The warp (exponential map) that is applied to the input image~\cref{eq:generalized-conv} can be implemented as follows. We start with an arbitrary pivot point $x_0$ in the image, and then sample all possible transformations of that point, $\{gx_0: g \in G\}$. For discrete images, $G$ will be implemented as a 2D grid of discrete transformations (e.g. rotations and scales at regular intervals), and $\{gx_0\}$ will be a 2D grid as well, referred to as the \emph{warp grid}. Finally, sampling the input image at the points $\{gx_0\}$ (for example, by bilinear interpolation) yields the warped image.

An illustration is given in \cref{fig:grids}, for various transformations (each one is discussed in more detail in \cref{sec:examples}). The red dot shows the pivot point $x_0$, and the two arrows pointing away from it show the original axes of the sampled grid of parameters. The grids were generated by sampling the transformation parameters at regular intervals. Note that the warp grids are independent of the image contents -- they can be computed once offline and then applied to any image.

The steps for implementing a warped convolution block are outlined in algorithm~1. 
The main advantage of implementing group convolution as warped convolution is that it replaces a large number of warping operations (one per group element) with a single warp.


\section{Discussion}

\subsection{Interpretation as a filtering operator in image space}
When $V \subset \mathbb{R}^2$, we can often interpret group convolution as an integration over image space, instead of over group elements. For this, we introduce the map $\eta(v)=\exp(v) x_0$ and assume that it is smooth, invertible, and surjective on the image domain $\Omega$. Surjectivity means that $G$ acts \emph{transitively} on $\Omega$, in the sense that every point $x\in\Omega$ can be reached from $x_0$ by a transformation $g$. Injectivity means that this transformation is unique.

Next, let  $h=\exp(v)$ and note that $\exp(v+u)=\exp(v)\exp(u)=h\exp(u)$. Hence
\begin{align*}
(\Ig \conv{G} \Fg)(h)
&= \int_V \Iw(v+u) \Fw(-u)\,du\\
&= \int_V I(h \exp(u)x_0) F(\exp(-u)x_0)\,du.
\end{align*}
Let the filter $F_-$ be the reflection\footnote{%
This is well defined because $\eta$ is invertible. In fact, if $x=gx_0=\exp(u)x_0=\eta(u)$, then $u = \eta^{-1}(x)$ and
$
F_-(x) 
= F(\eta(-\eta^{-1}(x))).
$
}
of $F$ around the pivot point $x_0$ by the group $G$, i.e.\
$
F_-(gx_0) = F(g^{-1}x_0). 
$
It follows that:
\begin{align*}
(\Ig \conv{G} \Fg)(h)
&= \int_V I(h \exp(u)x_0) F_-(\exp(u)x_0)\,du.
\end{align*}
We can now use the change of variable $u \leftarrow \eta^{-1}(x)$ to write
\begin{align}\label{eq:image-conv}
(\Ig \conv{G} \Fg)(h)
&= \int_V I(h \eta(u)) F_-(\eta(u))\,du\\
&= \int_\Omega I(h x) F_-(x) 
\left|
\frac{d\eta^{-1}}{dx}
\right| \,dx.
\end{align}
Thus we see that the group convolution amounts to applying a certain filter $F_-$ to the warped image $I \circ h$. The filter elements are reweighed by the determinant of the Jacobian of $\eta^{-1}$, which accounts for the stretching and shrinking of space due to the non-linear map. In practice, both the reflection and Jacobian can be absorbed into a learned filter, making such calculations unnecessary. Nevertheless, they offer a complementary view of warped convolutions.

\begin{algorithm}[t]\label{alg:main}

\vspace{0.7em}
\textit{Grid generation (offline)}
\begin{itemize}
    \item Compute the 2D warp grid $w=g_u(x_0)$ by applying a two-parameter spatial transformation $g: \R^2 \times \R^2 \rightarrow \R^2$ to a single pivot point $x_0$, using a 2D grid of parameters $u=\{(u_1+i\delta_1,u_2+j\delta_2): i=0,\ldots,m, j=0,\ldots,n\}$.
\end{itemize}

\textit{Warped convolution}
\begin{enumerate}  \setlength\itemsep{4pt} \setlength{\parskip}{0pt} \setlength{\parsep}{0pt}
\item Resample input image $I$ using the warp grid $w$, by bilinear interpolation.
\item Convolve warped image $\Iw$ with a learned filter $\Fw$.
\end{enumerate}

By \cref{eq:warped-conv}, and for appropriate transformations, these steps are equivalent to group convolution (which performs an exhaustive search across the pose-space of transformations), but at a much lower computational cost.

\caption{Warped Convolution}

\end{algorithm}

\begin{figure*}[!ht]\hspace*{\fill}
\resizebox{!}{0.23\textwidth}{\input{figures/warp-none.tikz}}\hspace{0.08cm}
\resizebox{!}{0.23\textwidth}{\input{figures/warp-hscale-vscale.tikz}}\hspace{0.08cm}
\resizebox{!}{0.23\textwidth}{\input{figures/warp-rotation-scale.tikz}}\hspace{0.08cm}
\resizebox{!}{0.23\textwidth}{\input{figures/warp-sphere.tikz}}\hspace*{\fill}\\[-0.3cm]

\hspace*{\fill}\includegraphics[width=0.95\textwidth]{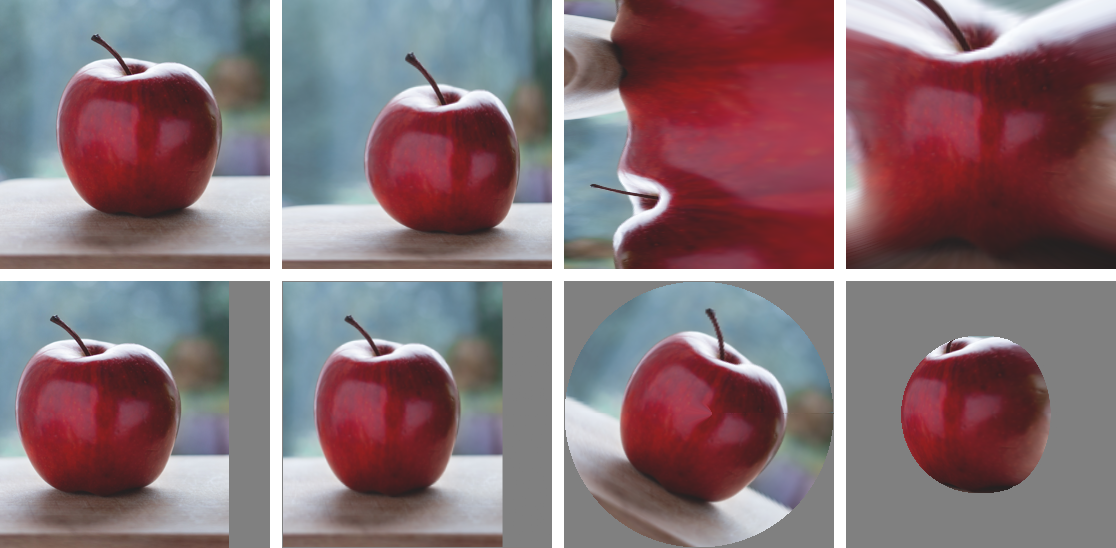}\hspace*{\fill}\\[-0.3cm]

\small\hspace*{\fill}
\begin{minipage}[t][1.5em]{0.23\textwidth}\center(a) translation\end{minipage}\hspace{0.08cm}
\begin{minipage}[t][1.5em]{0.23\textwidth}\center(b) scale/aspect ratio\end{minipage}\hspace{0.08cm}
\begin{minipage}[t][1.5em]{0.23\textwidth}\center(c) scale/rotation\end{minipage}\hspace{0.08cm}
\begin{minipage}[t][1.5em]{0.23\textwidth}\center(d) 3D rotation (yaw/pitch)\end{minipage}\hspace*{\fill}\\[-0.1cm]


\caption{First row: Sampling grids that define the warps associated with different spatial transformations. Second row: An example image (a) after warping with each grid (b-d). Third row: A small translation is applied to each warped image, which is then mapped back to the original space (by an inverse warp). Translation in one axis of the appropriate warped space is equivalent to (b) horizontal scaling; (c) planar rotation; (d) 3D rotation around the vertical axis.}
\label{fig:grids}
\end{figure*}

\subsection{Efficiency vs. generality}

By reducing to standard convolution, warped convolution allows one to take full advantage of modern convolution implementations~\cite{lavin2015fast,lyons2010understanding}, including those with lower computational complexity (e.g. FFT~\cite{lyons2010understanding}). However, while warped convolution works with an important class of transformations (including the ones considered in previous works~\citet{kanazawa2014locally,taco_cohen_learning_2014,marcos2016learning}), non-trivial restrictions are imposed on the transformation group: it must be Abelian and have only two parameters.

By contrast, the group-theoretic convolution operator of~\cref{eq:group-conv} does not make (almost) any restriction on the transformation group. Unfortunately, it is in general significantly more difficult to compute efficiently than the special case we consider here. To understand some of the implementation challenges, consider specializing~\cref{eq:image-space-group-conv} to a discrete group $G$ such as a discrete set of planar rotations. In this case the Haar measure is trivial and equal to 1, and one has:
$$
(I \conv{G} F)(t) = \sum_{g\in G} I(tgx_{0})F(g^{-1}x_{0}).
$$
Direct computation of this equation has complexity $\mathcal{O}(|G|^2)$ where $|G|$ is the cardinality of the discrete group. Assuming that $|G|$ is in the order of $\mathcal{O}(N^2)$ where $N$ is the resolution of the input image (as it would be for standard convolution), one would obtain a complexity of $\mathcal{O}(N^4)$. In practice, since usually the support of a $M\times M$ filter is much smaller than the image, this complexity might reduce to $\mathcal{O}(N^2M^2)$, which is the  complexity for the spatial domain implementation of convolution; however, compared to the standard case, this has two major disadvantages. First, the image is sampled in a spatially-varying manner, using bilinear or other interpolation, which foregoes the benefit of the regular, predictable, and local pattern of computations in standard convolution. This makes high-performance implementation of the naive algorithm difficult, particularly on GPUs. Secondly, it precludes the use of \emph{faster} convolution routines such as Winograd's algorithm~\cite{lavin2015fast} or the Fast Fourier Transform~\cite{lyons2010understanding}, the later having lower computational complexity than exhaustive search ($\mathcal{O}(N^{2} \log N)$).
The development of analogues of the FFT for other general groups remains the subject of active research~\cite{tygert2010fast,li2016interpolative}, which we sidestep by reusing highly optimized standard convolutions.



In practice, most recent works focus on very coarse transformations that do not change the filter support and can be implemented strictly via permutations, like horizontal/vertical flips and 90$^\circ$ rotations~\cite{dieleman2015rotation,taco_cohen_learning_2014}. Such difficulties may explain why group convolutions are not as widespread as CNNs.

\section{Examples of spatial transformations\label{sec:examples}}

We now give some concrete examples of two-parameter spatial transformations that obey the conditions of \cref{sub:calculations}, and can be useful in practice.

\subsection{Scale and aspect ratio}

Visual object detection tasks require predicting the extent of an object as a bounding box. While the location can be found accurately by a standard CNN, which is equivariant to translation, the size prediction could similarly benefit from equivariance to horizontal and vertical scale (equivalently, scale and aspect ratio).

Such a spatial transformation, from which a warp can be constructed, is given by:

\[
g_u(x)=\left[\begin{array}{c}
x_{1}s^{u_{1}}\\
x_{2}s^{u_{2}}
\end{array}\right]
\]

The $s$ constant controls the total degree of scaling applied. Notice that the output must be exponential in the scale parameters $u$; this ensures the additive structure of the exponential map ($\exp(v+u)=\exp(v)\exp(u)$). The resulting warp grid can be visualized in \cref{fig:grids}-b. In this case, the domain of the image must be $\Omega \in \R^2_+$, since a pivot $x_0$ in one quadrant cannot reach another quadrant by any amount of (positive) scaling.

\subsection{Scale and rotation (log-polar warp)\label{sub:scale-rot}}

Planar scale and rotation are perhaps the most obvious spatial transformations in images, and are a natural test case for works on spatial transformations \cite{kanazawa2014locally,marcos2016learning}. Rotating a point $x$ by $u_1$ radians and scaling it by $u_2$, around the origin, can be performed with

\vspace{-1em}

\[
g_u(x)=\left[\begin{array}{c}
s^{u_{2}}\left\Vert x\right\Vert \cos(\mathrm{atan}_2(x_{2},\,x_{1})+u_{1})\\
s^{u_{2}}\left\Vert x\right\Vert \sin(\mathrm{atan}_2(x_{2},\,x_{1})+u_{1})
\end{array}\right],
\]

\noindent where atan$_2$ is the standard 4-quadrant inverse tangent function (\texttt{atan2}). The domain in this case must exclude the origin ($\Omega \in \R^2 \setminus \{0\}$), since a pivot $x_0=0$ cannot reach any other points in the image by rotation or scaling.

The resulting warp grid can be visualized in \cref{fig:grids}-c. It is interesting to observe that it corresponds exactly to the log-polar domain, which is used in the signal processing literature to perform correlation across scale and rotation \cite{tzimiropoulos2010robust,reddy1996fft}. In fact, it was the source of inspiration for this work, which can be seen as a generalization of the log-polar domain to other spatial transformations.

\subsection{3D sphere rotation under perspective\label{sub:sphere}}

We will now tackle a more difficult spatial transformation, in an attempt to demonstrate the generality of our result. The transformations we will consider are yaw and pitch rotations in 3D space, as seen by a perspective camera. In the experiments (\cref{sec:experiments}) we will show how to apply it to face pose estimation.

In order to maintain additivity, the rotated 3D points must remain on the surface of a sphere. We consider a simplified camera and world model, whose only hyperparameters are a focal length $f$, the radius of a sphere $r$, and its distance from the camera center $d$. The equations for the spatial transformation corresponding to yaw and pitch rotation under this model are in \cref{sec:3d}.

The corresponding warp grid can be seen in \cref{fig:grids}-d. It can be observed that the grid corresponds to what we would expect of a 3D rendering of a sphere with a discrete mesh. An intuitive picture of the effect of the warp grid in such cases is that it wraps the 2D image around the surface of the 3D object, so that translation in the warped space corresponds to moving between vertexes of the 3D geometry.

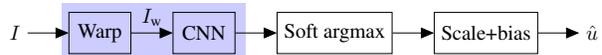
\begin{figure}
\centering
\resizebox{\columnwidth}{!}{%
\begin{tikzpicture}[auto, node distance=2cm, ampersand replacement=\&]

\tikzstyle{block} = [draw, rectangle, minimum height=2em, minimum width=3em]

\matrix(A)[column sep = 2em, row sep = 2em]
{
\node (I){$I$}; \& \node [block](warp){Warp}; \& \node [block](cnn){CNN}; \& \node [block](softargmax){Soft argmax}; \& \node [block](scalebias){Scale+bias}; \& \node (pred){$\hat{u}$}; \\
};

\draw [-triangle 45] (I) -- (warp);
\draw [-triangle 45] (warp) edge node {$\Iw$} (cnn);
\draw [-triangle 45] (cnn) -- (softargmax);
\draw [-triangle 45] (softargmax) -- (scalebias);
\draw [-triangle 45] (scalebias) -- (pred);

\begin{scope}[on background layer]
\fill[blue!20, transform canvas={xscale=1.1, yscale=1.4, xshift=0.75em}] (warp.west|-warp.north) rectangle (cnn.south-|cnn.east);
\end{scope}

\end{tikzpicture}
}%

\caption{Equivariant pose estimation strategy used in the experiments (\cref{sec:experiments}). With an appropriate warp and a standard CNN, the shaded block becomes equivalent to a group-equivariant CNN, which performs exhaustive searches across pose-space instead of image-space.}
\label{fig:arch}
\end{figure}

\section{Experiments\label{sec:experiments}}

\subsection{Architecture\label{sub:arch}}

As mentioned in \cref{sub:generalized-conv}, group convolution can perform an exhaustive search for patterns across spatial transformations, by varying pose parameters. For tasks where invariance to that transformation is important, it is usual to pool the detection responses across all poses \cite{marcos2016learning,kanazawa2014locally}.

In the experiments, however, we will test the framework in pose prediction tasks. As such, we do not want to pool the detection responses (e.g. with a max operation) but rather find the pose with the strongest response (i.e., an argmax operation). To perform this operation in a differentiable manner, we implement a soft argmax operation, defined as follows:

\vspace{-1em}

\begin{equation}\label{eq:softargmax}
s_{1}(a)=\sum_{ij}^{mn}\frac{i}{m}\sigma_{ij}(a),\qquad s_{2}(a)=\sum_{ij}^{mn}\frac{j}{n}\sigma_{ij}(a),
\end{equation}

\noindent where $\sigma(a) \in \R^{m \times n}$ is the softmax over all spatial locations, and $\sigma_{ij}(a)$ indexes the element at $(i,j)$. The outputs are the two spatial coordinates of the maximum value, $s(a) \in \R^2$.


Our base architecture then consists of the following blocks, outlined in \cref{fig:arch}. First, the input image is warped with a pre-generated grid, according to \cref{alg:main}. The warped image is then processed by a standard CNN, which is now equivariant to the  spatial transformation that was used to generate the warp grid. A soft argmax (\cref{eq:softargmax}) then finds the maximum over pose-space. To ensure the pose prediction is well registered to the reference coordinate system, a learnable scale and bias are applied to the outputs; multiple predictions can be linearly combined into a single one at this stage. Training proceeds by minimizing the $L^1$ loss between the predicted pose and ground truth pose.


%
%
%

\subsection{Google Earth\label{sub:earth}}

For the first task in our experiments, we will consider aerial photos of vehicles, which have been used in several works that deal with rotation invariance \cite{liu2014,schmidt_learning_2012,henriques2014pose}.

\paragraph{Dataset.} The Google Earth dataset \cite{heitz2008learning} contains bounding box annotations, supplemented with angle annotations from \cite{henriques2014pose}, for 697 vehicles in 15 large images. We use the first 10 for training and the rest for validation. Going beyond these previous works, we focus on the estimation of both rotation and scale parameters. The object scale is taken to be the diagonal length of the bounding box. Due to its small size, we augment the dataset during training, by randomly rotating the images uniformly over 360$^{\circ}$ and scaling them by up to 20\%.

\paragraph{Implementation.} A $48\times48$ image is cropped around each vehicle, and then fed to a network for pose prediction. The proposed method, Warped CNN, follows the architecture of \cref{sub:arch} (visualized in \cref{fig:arch}). The CNN block contains 3 convolutional layers with $3\times3$ filters, with 50, 20 and 50 output channels respectively. We use dilation factors of 2, 4 and 8 respectively (\emph{\`a trous} convolution~\cite{chen2015semantic}), which increases the receptive field and resolution without adding paramenters. There is a batch normalization and ReLU layer after each convolution, and a $3\times3$ max-pooling operator (stride 2) after the second one. The CNN block outputs response maps over 2D pose-space, which in this case consists of rotation and scale. All networks are trained for 40 epochs using the ADAM solver~\cite{kingma2015adam} and implemented in MatConvNet~\cite{vedaldi15matconvnet}. Angular error is taken modulo 180$^\circ$ due to annotation ambiguity. We report the average validation error over 3 runs.

\paragraph{Baselines and results.} The results of the experiments are presented in \cref{tab:earth}, which shows angular and scale error in the validation set. To verify whether the proposed warped convolution is indeed responsible for a boost in performance, rather than other architectural details, we compare it against a number of baselines with different components removed. The first baseline, CNN+softargmax, consists of the same architecture but without the warp (\cref{sub:scale-rot}). This is a standard CNN, with the soft argmax at the end. Since CNNs are equivariant to translation, rather than scale and rotation, we observe a drop in performance. For the second baseline, CNN+FC, we replace the soft argmax with a fully-connected (FC) layer, to allow a prediction that is not equivariant with translation. Due to the larger number of parameters, there is overfitting and a large drop in performance. We also compare against the method of \cite{dieleman2015rotation}, which applies the same CNN to 90$^\circ$ rotations and flips of the image, combining the result with a FC layer. Just like with CNN+FC, there is overfitting on this small dataset, which requires very fine angular predictions. The proposed Warped CNN achieves the best results, except for scale prediction where \cite{dieleman2015rotation} performs better. Our method has the same runtime performance as the CNN baselines, since the cost of a single warp is negligible, however \cite{dieleman2015rotation} is 16$\times$ slower, since it applies the same CNN to multiple transformed images.

\begin{table}
\caption{Results of scale and rotation pose estimation of vehicles in the Google Earth dataset (errors in pixels and degrees, resp.).}
\vskip 0.15in
\begin{center}
\begin{small}
\begin{sc}
\begin{tabular}{lcc}
\toprule
 & Rot. err. & Scale err. \tabularnewline
\midrule
CNN+FC & 22.54 & 5.04 \tabularnewline
CNN+softargmax & 9.36 & 4.87 \tabularnewline
Warped CNN & 8.29 & 4.79 \tabularnewline
\cite{dieleman2015rotation} & 31.11 & 4.29 \tabularnewline
\bottomrule
\end{tabular}
\end{sc}
\end{small}
\end{center}
\vskip -0.1in
\label{tab:earth}
\end{table}


\subsection{Faces\label{sub:faces}}

We now turn to face pose estimation in unconstrained photos, which requires handling more complex 3D rotations under perspective.

\paragraph{Dataset.} For this task we use the Annotated Facial Landmarks in the Wild (AFLW) dataset \cite{koestinger11b}. It contains about 25K faces found in Flickr photos, and includes yaw (left-right) and pitch (up-down) annotations. We removed 933 faces with yaw larger than 90 degrees (i.e., facing away from the camera), resulting in a set of 24,384 samples. 20\% of the faces were set aside for validation.

\paragraph{Implementation.} The region in each face's bounding box is resized to a $64\times64$ image, which is then processed by the network. Recall that our simplified 3D model of yaw and pitch rotation (\cref{sub:sphere}) assumes a spherical geometry. Although a person's head roughly follows a spherical shape, the sample images are centered around the face, not the head. As such, we use an affine Spatial Transformer Network (STN) \cite{jaderberg2015spatial} as a first step, to center the image correctly. Similarly, because the optimal camera parameters ($f$, $r$ and $d$) are difficult to set by hand, we let the network learn them, by computing their derivatives numerically (which has a low overhead, since they are scalars). The rest of the network follows the same diagram as before (\cref{fig:arch}). The main CNN has 4 convolutional layers, the first two with $5\times5$ filters, the others being $9\times9$. The numbers of output channels are 20, 50, 20 and 50, respectively. A $3\times3$ max-pooling with a stride of 2 is performed after the first layer, and there are ReLU non-linearities between the others. As for the STN, it has 3 convolutional layers ($5\times5$), with 20, 50 and 6 output channels respectively, and $3\times3$ max-pooling (stride 2) between them. The remaining experimental settings are as in \cref{sub:earth}.

\paragraph{Baselines and results.} The angular error of the proposed equivariant pose estimation, Warped CNN, is shown in \cref{tab:faces}, along with a number of baselines. The goal of these experiments is to demonstrate that it is possible to achieve equivariance to complex 3D rotations. To compare with non-equivariant models, we test two baselines with the same CNN architecture, where the softargmax is replaced with a fully-connected (FC) layer. We include the Spatial Transformer Network~\cite{jaderberg2015spatial} and CNN+FC, which is a standard CNN of equivalent (slightly larger) capacity. We observe that neither the FC or the STN components account for the performance of the warped convolution, which better exploits the natural 3D rotation equivariance of the data.

\begin{table}
\caption{Results of yaw and pitch pose estimation of faces on the AFLW dataset (error in degrees).}
\begin{center}
\begin{small}
\begin{sc}
\begin{tabular}{lcc}
\toprule 
 & Yaw err.  & Pitch err.\tabularnewline
\midrule
CNN+FC & 12.56 & 6.59 \tabularnewline
STN~\cite{jaderberg2015spatial}\hspace{-1em} & 13.65  & 7.22 \tabularnewline
Warped CNN & 7.07  & 5.28 \tabularnewline
\bottomrule
\end{tabular}
\end{sc}
\end{small}
\end{center}
\vskip -0.1in
\label{tab:faces}
\end{table}

\section{Conclusions}

In this work we show that it is possible to reuse highly optimized convolutional blocks, which are equivariant to image translation, and coax them to exhibit equivariance to other operators, including 3D transformations. This is achieved by a simple warp of the input image, implemented with off-the-shelf components of deep networks, and can be used for image recognition tasks involving a large range of image transformations.
Compared to other works, warped convolutions are simpler, relying on highly optimized convolution routines, and can flexibly handle many types of continuous transformations. Studying generalizations that support more than two parameters seems like a fruitful direction for future work.
In addition to the practical aspects, our analysis offers some insights into the fundamental relationships between arbitrary image transformations and convolutional architectures.



\appendix

\section{Spatial transformation for 3D sphere rotation under perspective} \label{sec:3d}

Our simplified model consists of a perspective camera with focal length $f$ and all other camera parameters equal to identity, at a distance $d$ from a centered sphere of radius $r$ (see \cref{fig:grids}-d).

A 2D point $x$ in image-space corresponds to the 3D point

\vspace{-1em}

\[
p=(x_{1},x_{2},f).
\]

\vspace{-0.5em}

Raycasting it along the $z$ axis, it will intersect the sphere surface at the 3D point

\vspace{-0.5em}

\[
q=\frac{p}{\left\Vert p\right\Vert }\left(k-\sqrt{k^{2}-d^{2}+r^{2}}\right),\,k=\frac{fd}{\left\Vert p\right\Vert }.
\]

\vspace{-0.5em}

If the argument of the square-root is negative, the ray does not intersect the sphere and so the point transformation is undefined. This means that the domain of the image $\Omega$ should be restricted to the sphere region. In practice, in such cases we simply leave the point unmodified. Then, the yaw and pitch coordinates of the point $q$ on the surface of the sphere are

\vspace{-1em}

\[
\phi_{1}=\cos^{-1}\left(-\frac{q_{2}}{r}\right),\,\phi_{2}=\mathrm{atan}_2\left(-\frac{q_{1}}{d-q_{3}}\right).
\]

\vspace{-0.5em}

These polar coordinates are now rotated by the spatial transformation parameters, $\phi'=\phi+u$. Converting them back to a 3D point $q'$


\vspace{-1.25em}

\[
q'=(r\sin\phi'_{1}\sin\phi'_{2}, -r\cos\phi'_{1}, r\sin\phi'_{1}\cos\phi'_{2}-d)
\]

\vspace{-0.25em}

Finally, projection of $q'$ into image-space yields

\vspace{-0.5em}


\[
g_u(x)=-\tfrac{f}{q'_{3}}\left(q'_{1}, q'_{2}\right).
\]

\vspace{-0.5em}

\paragraph{Acknowledgements.} This research was funded by ERC StG 638009-IDIU.

\bibliographystyle{icml2017}
\bibliography{references}

\vspace{-0.5em}

\end{document}

%% file: figures/warp-none.tikz
%
%
\definecolor{mycolor1}{rgb}{0.00000,0.44700,0.74100}%
\begin{tikzpicture}

\begin{axis}[%
width=3.642in,
height=3.642in,
at={(1.225in,0.492in)},
scale only axis,
xmin=-1.02,
xmax=1.02,
ymin=-1.02,
ymax=1.02,
hide axis,
axis x line*=bottom,
axis y line*=left
]
\addplot [color=mycolor1,solid,forget plot]
  table[row sep=crcr]{%
-1	-1\\
-1	-0.8\\
-1	-0.6\\
-1	-0.4\\
-1	-0.2\\
-1	0\\
-1	0.2\\
-1	0.4\\
-1	0.6\\
-1	0.8\\
-1	1\\
};
\addplot [color=mycolor1,solid,forget plot]
  table[row sep=crcr]{%
-0.8	-1\\
-0.8	-0.8\\
-0.8	-0.6\\
-0.8	-0.4\\
-0.8	-0.2\\
-0.8	0\\
-0.8	0.2\\
-0.8	0.4\\
-0.8	0.6\\
-0.8	0.8\\
-0.8	1\\
};
\addplot [color=mycolor1,solid,forget plot]
  table[row sep=crcr]{%
-0.6	-1\\
-0.6	-0.8\\
-0.6	-0.6\\
-0.6	-0.4\\
-0.6	-0.2\\
-0.6	0\\
-0.6	0.2\\
-0.6	0.4\\
-0.6	0.6\\
-0.6	0.8\\
-0.6	1\\
};
\addplot [color=mycolor1,solid,forget plot]
  table[row sep=crcr]{%
-0.4	-1\\
-0.4	-0.8\\
-0.4	-0.6\\
-0.4	-0.4\\
-0.4	-0.2\\
-0.4	0\\
-0.4	0.2\\
-0.4	0.4\\
-0.4	0.6\\
-0.4	0.8\\
-0.4	1\\
};
\addplot [color=mycolor1,solid,forget plot]
  table[row sep=crcr]{%
-0.2	-1\\
-0.2	-0.8\\
-0.2	-0.6\\
-0.2	-0.4\\
-0.2	-0.2\\
-0.2	0\\
-0.2	0.2\\
-0.2	0.4\\
-0.2	0.6\\
-0.2	0.8\\
-0.2	1\\
};
\addplot [color=mycolor1,solid,forget plot]
  table[row sep=crcr]{%
0	-1\\
0	-0.8\\
0	-0.6\\
0	-0.4\\
0	-0.2\\
0	0\\
0	0.2\\
0	0.4\\
0	0.6\\
0	0.8\\
0	1\\
};
\addplot [color=mycolor1,solid,forget plot]
  table[row sep=crcr]{%
0.2	-1\\
0.2	-0.8\\
0.2	-0.6\\
0.2	-0.4\\
0.2	-0.2\\
0.2	0\\
0.2	0.2\\
0.2	0.4\\
0.2	0.6\\
0.2	0.8\\
0.2	1\\
};
\addplot [color=mycolor1,solid,forget plot]
  table[row sep=crcr]{%
0.4	-1\\
0.4	-0.8\\
0.4	-0.6\\
0.4	-0.4\\
0.4	-0.2\\
0.4	0\\
0.4	0.2\\
0.4	0.4\\
0.4	0.6\\
0.4	0.8\\
0.4	1\\
};
\addplot [color=mycolor1,solid,forget plot]
  table[row sep=crcr]{%
0.6	-1\\
0.6	-0.8\\
0.6	-0.6\\
0.6	-0.4\\
0.6	-0.2\\
0.6	0\\
0.6	0.2\\
0.6	0.4\\
0.6	0.6\\
0.6	0.8\\
0.6	1\\
};
\addplot [color=mycolor1,solid,forget plot]
  table[row sep=crcr]{%
0.8	-1\\
0.8	-0.8\\
0.8	-0.6\\
0.8	-0.4\\
0.8	-0.2\\
0.8	0\\
0.8	0.2\\
0.8	0.4\\
0.8	0.6\\
0.8	0.8\\
0.8	1\\
};
\addplot [color=mycolor1,solid,forget plot]
  table[row sep=crcr]{%
1	-1\\
1	-0.8\\
1	-0.6\\
1	-0.4\\
1	-0.2\\
1	0\\
1	0.2\\
1	0.4\\
1	0.6\\
1	0.8\\
1	1\\
};
\addplot [color=mycolor1,solid,forget plot]
  table[row sep=crcr]{%
-1	-1\\
-0.8	-1\\
-0.6	-1\\
-0.4	-1\\
-0.2	-1\\
0	-1\\
0.2	-1\\
0.4	-1\\
0.6	-1\\
0.8	-1\\
1	-1\\
};
\addplot [color=mycolor1,solid,forget plot]
  table[row sep=crcr]{%
-1	-0.8\\
-0.8	-0.8\\
-0.6	-0.8\\
-0.4	-0.8\\
-0.2	-0.8\\
0	-0.8\\
0.2	-0.8\\
0.4	-0.8\\
0.6	-0.8\\
0.8	-0.8\\
1	-0.8\\
};
\addplot [color=mycolor1,solid,forget plot]
  table[row sep=crcr]{%
-1	-0.6\\
-0.8	-0.6\\
-0.6	-0.6\\
-0.4	-0.6\\
-0.2	-0.6\\
0	-0.6\\
0.2	-0.6\\
0.4	-0.6\\
0.6	-0.6\\
0.8	-0.6\\
1	-0.6\\
};
\addplot [color=mycolor1,solid,forget plot]
  table[row sep=crcr]{%
-1	-0.4\\
-0.8	-0.4\\
-0.6	-0.4\\
-0.4	-0.4\\
-0.2	-0.4\\
0	-0.4\\
0.2	-0.4\\
0.4	-0.4\\
0.6	-0.4\\
0.8	-0.4\\
1	-0.4\\
};
\addplot [color=mycolor1,solid,forget plot]
  table[row sep=crcr]{%
-1	-0.2\\
-0.8	-0.2\\
-0.6	-0.2\\
-0.4	-0.2\\
-0.2	-0.2\\
0	-0.2\\
0.2	-0.2\\
0.4	-0.2\\
0.6	-0.2\\
0.8	-0.2\\
1	-0.2\\
};
\addplot [color=mycolor1,solid,forget plot]
  table[row sep=crcr]{%
-1	0\\
-0.8	0\\
-0.6	0\\
-0.4	0\\
-0.2	0\\
0	0\\
0.2	0\\
0.4	0\\
0.6	0\\
0.8	0\\
1	0\\
};
\addplot [color=mycolor1,solid,forget plot]
  table[row sep=crcr]{%
-1	0.2\\
-0.8	0.2\\
-0.6	0.2\\
-0.4	0.2\\
-0.2	0.2\\
0	0.2\\
0.2	0.2\\
0.4	0.2\\
0.6	0.2\\
0.8	0.2\\
1	0.2\\
};
\addplot [color=mycolor1,solid,forget plot]
  table[row sep=crcr]{%
-1	0.4\\
-0.8	0.4\\
-0.6	0.4\\
-0.4	0.4\\
-0.2	0.4\\
0	0.4\\
0.2	0.4\\
0.4	0.4\\
0.6	0.4\\
0.8	0.4\\
1	0.4\\
};
\addplot [color=mycolor1,solid,forget plot]
  table[row sep=crcr]{%
-1	0.6\\
-0.8	0.6\\
-0.6	0.6\\
-0.4	0.6\\
-0.2	0.6\\
0	0.6\\
0.2	0.6\\
0.4	0.6\\
0.6	0.6\\
0.8	0.6\\
1	0.6\\
};
\addplot [color=mycolor1,solid,forget plot]
  table[row sep=crcr]{%
-1	0.8\\
-0.8	0.8\\
-0.6	0.8\\
-0.4	0.8\\
-0.2	0.8\\
0	0.8\\
0.2	0.8\\
0.4	0.8\\
0.6	0.8\\
0.8	0.8\\
1	0.8\\
};
\addplot [color=mycolor1,solid,forget plot]
  table[row sep=crcr]{%
-1	1\\
-0.8	1\\
-0.6	1\\
-0.4	1\\
-0.2	1\\
0	1\\
0.2	1\\
0.4	1\\
0.6	1\\
0.8	1\\
1	1\\
};
\addplot [color=red,mark size=3.3pt,only marks,mark=*,mark options={solid},forget plot]
  table[row sep=crcr]{%
-1	-1\\
};

\addplot[area legend,solid,line width=1.5pt,draw=black,fill=black,forget plot]
table[row sep=crcr] {%
x	y\\
-1	-0.926765690721012\\
-1	-1\\
-1	-1\\
-1	-1\\
-1	-1\\
-1	-1\\
-1	-0.853531381442023\\
-1.02082001139853	-0.857202511190241\\
-1	-0.8\\
-0.979179988601473	-0.857202511190241\\
-1	-0.853531381442023\\
}--cycle;

\addplot[area legend,solid,line width=1.5pt,draw=black,fill=black,forget plot]
table[row sep=crcr] {%
x	y\\
-0.933457113401265	-1\\
-1	-1\\
-1	-1\\
-1	-1\\
-1	-1\\
-1	-1\\
-0.866914226802529	-1\\
-0.871503138987802	-1.02082001139853\\
-0.8	-1\\
-0.871503138987802	-0.979179988601473\\
-0.866914226802529	-1\\
}--cycle;
\end{axis}
\end{tikzpicture}%

%% file: figures/warp-hscale-vscale.tikz
%
%
\definecolor{mycolor1}{rgb}{0.00000,0.44700,0.74100}%
\begin{tikzpicture}

\begin{axis}[%
width=3.566in,
height=3.566in,
at={(1.236in,0.481in)},
scale only axis,
xmin=-0.965872168540955,
xmax=0.879403471946716,
ymin=-0.965872168540955,
ymax=0.879403471946716,
hide axis,
axis x line*=bottom,
axis y line*=left
]
\addplot [color=mycolor1,solid,forget plot]
  table[row sep=crcr]{%
-0.945872187614441	-0.945872187614441\\
-0.945872187614441	-0.8299480676651\\
-0.945872187614441	-0.702991545200348\\
-0.945872187614441	-0.563952565193176\\
-0.945872187614441	-0.411681413650513\\
-0.945872187614441	-0.244918659329414\\
-0.945872187614441	-0.0622851997613907\\
-0.945872187614441	0.13772939145565\\
-0.945872187614441	0.356779247522354\\
-0.945872187614441	0.596675992012024\\
-0.945872187614441	0.859403491020203\\
};
\addplot [color=mycolor1,solid,forget plot]
  table[row sep=crcr]{%
-0.8299480676651	-0.945872187614441\\
-0.8299480676651	-0.8299480676651\\
-0.8299480676651	-0.702991545200348\\
-0.8299480676651	-0.563952565193176\\
-0.8299480676651	-0.411681413650513\\
-0.8299480676651	-0.244918659329414\\
-0.8299480676651	-0.0622851997613907\\
-0.8299480676651	0.13772939145565\\
-0.8299480676651	0.356779247522354\\
-0.8299480676651	0.596675992012024\\
-0.8299480676651	0.859403491020203\\
};
\addplot [color=mycolor1,solid,forget plot]
  table[row sep=crcr]{%
-0.702991545200348	-0.945872187614441\\
-0.702991545200348	-0.8299480676651\\
-0.702991545200348	-0.702991545200348\\
-0.702991545200348	-0.563952565193176\\
-0.702991545200348	-0.411681413650513\\
-0.702991545200348	-0.244918659329414\\
-0.702991545200348	-0.0622851997613907\\
-0.702991545200348	0.13772939145565\\
-0.702991545200348	0.356779247522354\\
-0.702991545200348	0.596675992012024\\
-0.702991545200348	0.859403491020203\\
};
\addplot [color=mycolor1,solid,forget plot]
  table[row sep=crcr]{%
-0.563952565193176	-0.945872187614441\\
-0.563952565193176	-0.8299480676651\\
-0.563952565193176	-0.702991545200348\\
-0.563952565193176	-0.563952565193176\\
-0.563952565193176	-0.411681413650513\\
-0.563952565193176	-0.244918659329414\\
-0.563952565193176	-0.0622851997613907\\
-0.563952565193176	0.13772939145565\\
-0.563952565193176	0.356779247522354\\
-0.563952565193176	0.596675992012024\\
-0.563952565193176	0.859403491020203\\
};
\addplot [color=mycolor1,solid,forget plot]
  table[row sep=crcr]{%
-0.411681413650513	-0.945872187614441\\
-0.411681413650513	-0.8299480676651\\
-0.411681413650513	-0.702991545200348\\
-0.411681413650513	-0.563952565193176\\
-0.411681413650513	-0.411681413650513\\
-0.411681413650513	-0.244918659329414\\
-0.411681413650513	-0.0622851997613907\\
-0.411681413650513	0.13772939145565\\
-0.411681413650513	0.356779247522354\\
-0.411681413650513	0.596675992012024\\
-0.411681413650513	0.859403491020203\\
};
\addplot [color=mycolor1,solid,forget plot]
  table[row sep=crcr]{%
-0.244918659329414	-0.945872187614441\\
-0.244918659329414	-0.8299480676651\\
-0.244918659329414	-0.702991545200348\\
-0.244918659329414	-0.563952565193176\\
-0.244918659329414	-0.411681413650513\\
-0.244918659329414	-0.244918659329414\\
-0.244918659329414	-0.0622851997613907\\
-0.244918659329414	0.13772939145565\\
-0.244918659329414	0.356779247522354\\
-0.244918659329414	0.596675992012024\\
-0.244918659329414	0.859403491020203\\
};
\addplot [color=mycolor1,solid,forget plot]
  table[row sep=crcr]{%
-0.0622851997613907	-0.945872187614441\\
-0.0622851997613907	-0.8299480676651\\
-0.0622851997613907	-0.702991545200348\\
-0.0622851997613907	-0.563952565193176\\
-0.0622851997613907	-0.411681413650513\\
-0.0622851997613907	-0.244918659329414\\
-0.0622851997613907	-0.0622851997613907\\
-0.0622851997613907	0.13772939145565\\
-0.0622851997613907	0.356779247522354\\
-0.0622851997613907	0.596675992012024\\
-0.0622851997613907	0.859403491020203\\
};
\addplot [color=mycolor1,solid,forget plot]
  table[row sep=crcr]{%
0.13772939145565	-0.945872187614441\\
0.13772939145565	-0.8299480676651\\
0.13772939145565	-0.702991545200348\\
0.13772939145565	-0.563952565193176\\
0.13772939145565	-0.411681413650513\\
0.13772939145565	-0.244918659329414\\
0.13772939145565	-0.0622851997613907\\
0.13772939145565	0.13772939145565\\
0.13772939145565	0.356779247522354\\
0.13772939145565	0.596675992012024\\
0.13772939145565	0.859403491020203\\
};
\addplot [color=mycolor1,solid,forget plot]
  table[row sep=crcr]{%
0.356779247522354	-0.945872187614441\\
0.356779247522354	-0.8299480676651\\
0.356779247522354	-0.702991545200348\\
0.356779247522354	-0.563952565193176\\
0.356779247522354	-0.411681413650513\\
0.356779247522354	-0.244918659329414\\
0.356779247522354	-0.0622851997613907\\
0.356779247522354	0.13772939145565\\
0.356779247522354	0.356779247522354\\
0.356779247522354	0.596675992012024\\
0.356779247522354	0.859403491020203\\
};
\addplot [color=mycolor1,solid,forget plot]
  table[row sep=crcr]{%
0.596675992012024	-0.945872187614441\\
0.596675992012024	-0.8299480676651\\
0.596675992012024	-0.702991545200348\\
0.596675992012024	-0.563952565193176\\
0.596675992012024	-0.411681413650513\\
0.596675992012024	-0.244918659329414\\
0.596675992012024	-0.0622851997613907\\
0.596675992012024	0.13772939145565\\
0.596675992012024	0.356779247522354\\
0.596675992012024	0.596675992012024\\
0.596675992012024	0.859403491020203\\
};
\addplot [color=mycolor1,solid,forget plot]
  table[row sep=crcr]{%
0.859403491020203	-0.945872187614441\\
0.859403491020203	-0.8299480676651\\
0.859403491020203	-0.702991545200348\\
0.859403491020203	-0.563952565193176\\
0.859403491020203	-0.411681413650513\\
0.859403491020203	-0.244918659329414\\
0.859403491020203	-0.0622851997613907\\
0.859403491020203	0.13772939145565\\
0.859403491020203	0.356779247522354\\
0.859403491020203	0.596675992012024\\
0.859403491020203	0.859403491020203\\
};
\addplot [color=mycolor1,solid,forget plot]
  table[row sep=crcr]{%
-0.945872187614441	-0.945872187614441\\
-0.8299480676651	-0.945872187614441\\
-0.702991545200348	-0.945872187614441\\
-0.563952565193176	-0.945872187614441\\
-0.411681413650513	-0.945872187614441\\
-0.244918659329414	-0.945872187614441\\
-0.0622851997613907	-0.945872187614441\\
0.13772939145565	-0.945872187614441\\
0.356779247522354	-0.945872187614441\\
0.596675992012024	-0.945872187614441\\
0.859403491020203	-0.945872187614441\\
};
\addplot [color=mycolor1,solid,forget plot]
  table[row sep=crcr]{%
-0.945872187614441	-0.8299480676651\\
-0.8299480676651	-0.8299480676651\\
-0.702991545200348	-0.8299480676651\\
-0.563952565193176	-0.8299480676651\\
-0.411681413650513	-0.8299480676651\\
-0.244918659329414	-0.8299480676651\\
-0.0622851997613907	-0.8299480676651\\
0.13772939145565	-0.8299480676651\\
0.356779247522354	-0.8299480676651\\
0.596675992012024	-0.8299480676651\\
0.859403491020203	-0.8299480676651\\
};
\addplot [color=mycolor1,solid,forget plot]
  table[row sep=crcr]{%
-0.945872187614441	-0.702991545200348\\
-0.8299480676651	-0.702991545200348\\
-0.702991545200348	-0.702991545200348\\
-0.563952565193176	-0.702991545200348\\
-0.411681413650513	-0.702991545200348\\
-0.244918659329414	-0.702991545200348\\
-0.0622851997613907	-0.702991545200348\\
0.13772939145565	-0.702991545200348\\
0.356779247522354	-0.702991545200348\\
0.596675992012024	-0.702991545200348\\
0.859403491020203	-0.702991545200348\\
};
\addplot [color=mycolor1,solid,forget plot]
  table[row sep=crcr]{%
-0.945872187614441	-0.563952565193176\\
-0.8299480676651	-0.563952565193176\\
-0.702991545200348	-0.563952565193176\\
-0.563952565193176	-0.563952565193176\\
-0.411681413650513	-0.563952565193176\\
-0.244918659329414	-0.563952565193176\\
-0.0622851997613907	-0.563952565193176\\
0.13772939145565	-0.563952565193176\\
0.356779247522354	-0.563952565193176\\
0.596675992012024	-0.563952565193176\\
0.859403491020203	-0.563952565193176\\
};
\addplot [color=mycolor1,solid,forget plot]
  table[row sep=crcr]{%
-0.945872187614441	-0.411681413650513\\
-0.8299480676651	-0.411681413650513\\
-0.702991545200348	-0.411681413650513\\
-0.563952565193176	-0.411681413650513\\
-0.411681413650513	-0.411681413650513\\
-0.244918659329414	-0.411681413650513\\
-0.0622851997613907	-0.411681413650513\\
0.13772939145565	-0.411681413650513\\
0.356779247522354	-0.411681413650513\\
0.596675992012024	-0.411681413650513\\
0.859403491020203	-0.411681413650513\\
};
\addplot [color=mycolor1,solid,forget plot]
  table[row sep=crcr]{%
-0.945872187614441	-0.244918659329414\\
-0.8299480676651	-0.244918659329414\\
-0.702991545200348	-0.244918659329414\\
-0.563952565193176	-0.244918659329414\\
-0.411681413650513	-0.244918659329414\\
-0.244918659329414	-0.244918659329414\\
-0.0622851997613907	-0.244918659329414\\
0.13772939145565	-0.244918659329414\\
0.356779247522354	-0.244918659329414\\
0.596675992012024	-0.244918659329414\\
0.859403491020203	-0.244918659329414\\
};
\addplot [color=mycolor1,solid,forget plot]
  table[row sep=crcr]{%
-0.945872187614441	-0.0622851997613907\\
-0.8299480676651	-0.0622851997613907\\
-0.702991545200348	-0.0622851997613907\\
-0.563952565193176	-0.0622851997613907\\
-0.411681413650513	-0.0622851997613907\\
-0.244918659329414	-0.0622851997613907\\
-0.0622851997613907	-0.0622851997613907\\
0.13772939145565	-0.0622851997613907\\
0.356779247522354	-0.0622851997613907\\
0.596675992012024	-0.0622851997613907\\
0.859403491020203	-0.0622851997613907\\
};
\addplot [color=mycolor1,solid,forget plot]
  table[row sep=crcr]{%
-0.945872187614441	0.13772939145565\\
-0.8299480676651	0.13772939145565\\
-0.702991545200348	0.13772939145565\\
-0.563952565193176	0.13772939145565\\
-0.411681413650513	0.13772939145565\\
-0.244918659329414	0.13772939145565\\
-0.0622851997613907	0.13772939145565\\
0.13772939145565	0.13772939145565\\
0.356779247522354	0.13772939145565\\
0.596675992012024	0.13772939145565\\
0.859403491020203	0.13772939145565\\
};
\addplot [color=mycolor1,solid,forget plot]
  table[row sep=crcr]{%
-0.945872187614441	0.356779247522354\\
-0.8299480676651	0.356779247522354\\
-0.702991545200348	0.356779247522354\\
-0.563952565193176	0.356779247522354\\
-0.411681413650513	0.356779247522354\\
-0.244918659329414	0.356779247522354\\
-0.0622851997613907	0.356779247522354\\
0.13772939145565	0.356779247522354\\
0.356779247522354	0.356779247522354\\
0.596675992012024	0.356779247522354\\
0.859403491020203	0.356779247522354\\
};
\addplot [color=mycolor1,solid,forget plot]
  table[row sep=crcr]{%
-0.945872187614441	0.596675992012024\\
-0.8299480676651	0.596675992012024\\
-0.702991545200348	0.596675992012024\\
-0.563952565193176	0.596675992012024\\
-0.411681413650513	0.596675992012024\\
-0.244918659329414	0.596675992012024\\
-0.0622851997613907	0.596675992012024\\
0.13772939145565	0.596675992012024\\
0.356779247522354	0.596675992012024\\
0.596675992012024	0.596675992012024\\
0.859403491020203	0.596675992012024\\
};
\addplot [color=mycolor1,solid,forget plot]
  table[row sep=crcr]{%
-0.945872187614441	0.859403491020203\\
-0.8299480676651	0.859403491020203\\
-0.702991545200348	0.859403491020203\\
-0.563952565193176	0.859403491020203\\
-0.411681413650513	0.859403491020203\\
-0.244918659329414	0.859403491020203\\
-0.0622851997613907	0.859403491020203\\
0.13772939145565	0.859403491020203\\
0.356779247522354	0.859403491020203\\
0.596675992012024	0.859403491020203\\
0.859403491020203	0.859403491020203\\
};
\addplot [color=red,mark size=3.3pt,only marks,mark=*,mark options={solid},forget plot]
  table[row sep=crcr]{%
-0.945872187614441	-0.945872187614441\\
};

\addplot[area legend,solid,line width=1.5pt,draw=black,fill=black,forget plot]
table[row sep=crcr] {%
x	y\\
-0.945872187614441	-0.915249347686768\\
-0.945872187614441	-0.945872187614441\\
-0.945872187614441	-0.945872187614441\\
-0.945872187614441	-0.945872187614441\\
-0.945872187614441	-0.945872187614441\\
-0.945872187614441	-0.945872187614441\\
-0.945872187614441	-0.884626567363739\\
-0.967138350009918	-0.888376355171204\\
-0.945872187614441	-0.8299480676651\\
-0.924606025218964	-0.888376355171204\\
-0.945872187614441	-0.884626567363739\\
}--cycle;

\addplot[area legend,solid,line width=1.5pt,draw=black,fill=black,forget plot]
table[row sep=crcr] {%
x	y\\
-0.915249347686768	-0.945872187614441\\
-0.945872187614441	-0.945872187614441\\
-0.945872187614441	-0.945872187614441\\
-0.945872187614441	-0.945872187614441\\
-0.945872187614441	-0.945872187614441\\
-0.945872187614441	-0.945872187614441\\
-0.884626567363739	-0.945872187614441\\
-0.888376355171204	-0.967138350009918\\
-0.8299480676651	-0.945872187614441\\
-0.888376355171204	-0.924606025218964\\
-0.884626567363739	-0.945872187614441\\
}--cycle;
\end{axis}
\end{tikzpicture}%

%% file: figures/warp-rotation-scale.tikz
%
%
\definecolor{mycolor1}{rgb}{0.00000,0.44700,0.74100}%
\begin{tikzpicture}

\begin{axis}[%
width=3.566in,
height=3.566in,
at={(1.236in,0.481in)},
scale only axis,
xmin=-0.979492962360382,
xmax=1.01999998092651,
ymin=-1.00982141494751,
ymax=1.00982141494751,
hide axis,
axis x line*=bottom,
axis y line*=left
]
\addplot [color=mycolor1,solid,forget plot]
  table[row sep=crcr]{%
1	0\\
0.841253519058228	0.540640830993652\\
0.415415018796921	0.909631967544556\\
-0.142314836382866	0.989821434020996\\
-0.654860734939575	0.755749583244324\\
-0.959492981433868	0.281732559204102\\
-0.959492981433868	-0.281732559204102\\
-0.654860734939575	-0.755749583244324\\
-0.142314836382866	-0.989821434020996\\
0.415415018796921	-0.909631967544556\\
0.841253519058228	-0.540640830993652\\
1	-1.11022302462516e-15\\
};
\addplot [color=mycolor1,solid,forget plot]
  table[row sep=crcr]{%
0.794328212738037	0\\
0.668231427669525	0.429446280002594\\
0.329975873231888	0.722546398639679\\
-0.113044694066048	0.786243140697479\\
-0.520174384117126	0.600313246250153\\
-0.762152373790741	0.223788127303123\\
-0.762152373790741	-0.223788127303123\\
-0.520174384117126	-0.600313246250153\\
-0.113044694066048	-0.786243140697479\\
0.329975873231888	-0.722546398639679\\
0.668231427669525	-0.429446280002594\\
0.794328212738037	-8.88178419700125e-16\\
};
\addplot [color=mycolor1,solid,forget plot]
  table[row sep=crcr]{%
0.630957365036011	0\\
0.530795097351074	0.341121286153793\\
0.262109160423279	0.573938965797424\\
-0.0897945910692215	0.624535083770752\\
-0.413189202547073	0.476845741271973\\
-0.605399131774902	0.177761226892471\\
-0.605399131774902	-0.177761226892471\\
-0.413189202547073	-0.476845741271973\\
-0.0897945910692215	-0.624535083770752\\
0.262109160423279	-0.573938965797424\\
0.530795097351074	-0.341121286153793\\
0.630957365036011	-6.66133814775094e-16\\
};
\addplot [color=mycolor1,solid,forget plot]
  table[row sep=crcr]{%
0.501187205314636	0\\
0.421625524759293	0.27096226811409\\
0.208200708031654	0.455895930528641\\
-0.07132638245821	0.49608588218689\\
-0.328207850456238	0.378772050142288\\
-0.480885624885559	0.141200765967369\\
-0.480885624885559	-0.141200765967369\\
-0.328207850456238	-0.378772050142288\\
-0.07132638245821	-0.49608588218689\\
0.208200708031654	-0.455895930528641\\
0.421625524759293	-0.27096226811409\\
0.501187205314636	-5.55111512312578e-16\\
};
\addplot [color=mycolor1,solid,forget plot]
  table[row sep=crcr]{%
0.398107171058655	0\\
0.334909051656723	0.215232983231544\\
0.16537968814373	0.362131029367447\\
-0.0566565580666065	0.394055008888245\\
-0.260704755783081	0.300869315862656\\
-0.381981045007706	0.112159751355648\\
-0.381981045007706	-0.112159751355648\\
-0.260704755783081	-0.300869315862656\\
-0.0566565580666065	-0.394055008888245\\
0.16537968814373	-0.362131029367447\\
0.334909051656723	-0.215232983231544\\
0.398107171058655	-4.44089209850063e-16\\
};
\addplot [color=mycolor1,solid,forget plot]
  table[row sep=crcr]{%
0.31622776389122	0\\
0.266027718782425	0.170965641736984\\
0.131365761160851	0.287650883197784\\
-0.0450039021670818	0.313009023666382\\
-0.207085147500038	0.238988995552063\\
-0.303418308496475	0.0890916585922241\\
-0.303418308496475	-0.0890916585922241\\
-0.207085147500038	-0.238988995552063\\
-0.0450039021670818	-0.313009023666382\\
0.131365761160851	-0.287650883197784\\
0.266027718782425	-0.170965641736984\\
0.31622776389122	-3.33066907387547e-16\\
};
\addplot [color=mycolor1,solid,forget plot]
  table[row sep=crcr]{%
0.251188635826111	0\\
0.211313337087631	0.135802835226059\\
0.104347534477711	0.228489220142365\\
-0.0357478708028793	0.248631909489632\\
-0.164493575692177	0.189835712313652\\
-0.241013735532761	0.0707680210471153\\
-0.241013735532761	-0.0707680210471153\\
-0.164493575692177	-0.189835712313652\\
-0.0357478708028793	-0.248631909489632\\
0.104347534477711	-0.228489220142365\\
0.211313337087631	-0.135802835226059\\
0.251188635826111	-3.33066907387547e-16\\
};
\addplot [color=mycolor1,solid,forget plot]
  table[row sep=crcr]{%
0.199526235461235	0\\
0.167852148413658	0.107872024178505\\
0.0828861892223358	0.181495442986488\\
-0.0283955428749323	0.197495341300964\\
-0.130661889910698	0.150791868567467\\
-0.191444024443626	0.0562130361795425\\
-0.191444024443626	-0.0562130361795425\\
-0.130661889910698	-0.150791868567467\\
-0.0283955428749323	-0.197495341300964\\
0.0828861892223358	-0.181495442986488\\
0.167852148413658	-0.107872024178505\\
0.199526235461235	-2.22044604925031e-16\\
};
\addplot [color=mycolor1,solid,forget plot]
  table[row sep=crcr]{%
0.158489316701889	0\\
0.133329704403877	0.0856857970356941\\
0.0658388435840607	0.144166961312294\\
-0.0225553810596466	0.156876131892204\\
-0.103788435459137	0.119778238236904\\
-0.152069389820099	0.0446516014635563\\
-0.152069389820099	-0.0446516014635563\\
-0.103788435459137	-0.119778238236904\\
-0.0225553810596466	-0.156876131892204\\
0.0658388435840607	-0.144166961312294\\
0.133329704403877	-0.0856857970356941\\
0.158489316701889	-2.22044604925031e-16\\
};
\addplot [color=mycolor1,solid,forget plot]
  table[row sep=crcr]{%
0.125892534852028	0\\
0.105907544493675	0.0680626481771469\\
0.0522976517677307	0.114515885710716\\
-0.0179163757711649	0.124611139297485\\
-0.082442082464695	0.095143236219883\\
-0.120793007314205	0.0354680269956589\\
-0.120793007314205	-0.0354680269956589\\
-0.082442082464695	-0.095143236219883\\
-0.0179163757711649	-0.124611139297485\\
0.0522976517677307	-0.114515885710716\\
0.105907544493675	-0.0680626481771469\\
0.125892534852028	-1.11022302462516e-16\\
};
\addplot [color=mycolor1,solid,forget plot]
  table[row sep=crcr]{%
0.100000001490116	0\\
0.084125354886055	0.054064080119133\\
0.0415415018796921	0.0909631997346878\\
-0.0142314834520221	0.098982147872448\\
-0.0654860734939575	0.0755749568343163\\
-0.095949299633503	0.0281732548028231\\
-0.095949299633503	-0.0281732548028231\\
-0.0654860734939575	-0.0755749568343163\\
-0.0142314834520221	-0.098982147872448\\
0.0415415018796921	-0.0909631997346878\\
0.084125354886055	-0.054064080119133\\
0.100000001490116	-1.11022302462516e-16\\
};
\addplot [color=mycolor1,solid,forget plot]
  table[row sep=crcr]{%
1	0\\
0.794328212738037	0\\
0.630957365036011	0\\
0.501187205314636	0\\
0.398107171058655	0\\
0.31622776389122	0\\
0.251188635826111	0\\
0.199526235461235	0\\
0.158489316701889	0\\
0.125892534852028	0\\
0.100000001490116	0\\
};
\addplot [color=mycolor1,solid,forget plot]
  table[row sep=crcr]{%
0.841253519058228	0.540640830993652\\
0.668231427669525	0.429446280002594\\
0.530795097351074	0.341121286153793\\
0.421625524759293	0.27096226811409\\
0.334909051656723	0.215232983231544\\
0.266027718782425	0.170965641736984\\
0.211313337087631	0.135802835226059\\
0.167852148413658	0.107872024178505\\
0.133329704403877	0.0856857970356941\\
0.105907544493675	0.0680626481771469\\
0.084125354886055	0.054064080119133\\
};
\addplot [color=mycolor1,solid,forget plot]
  table[row sep=crcr]{%
0.415415018796921	0.909631967544556\\
0.329975873231888	0.722546398639679\\
0.262109160423279	0.573938965797424\\
0.208200708031654	0.455895930528641\\
0.16537968814373	0.362131029367447\\
0.131365761160851	0.287650883197784\\
0.104347534477711	0.228489220142365\\
0.0828861892223358	0.181495442986488\\
0.0658388435840607	0.144166961312294\\
0.0522976517677307	0.114515885710716\\
0.0415415018796921	0.0909631997346878\\
};
\addplot [color=mycolor1,solid,forget plot]
  table[row sep=crcr]{%
-0.142314836382866	0.989821434020996\\
-0.113044694066048	0.786243140697479\\
-0.0897945910692215	0.624535083770752\\
-0.07132638245821	0.49608588218689\\
-0.0566565580666065	0.394055008888245\\
-0.0450039021670818	0.313009023666382\\
-0.0357478708028793	0.248631909489632\\
-0.0283955428749323	0.197495341300964\\
-0.0225553810596466	0.156876131892204\\
-0.0179163757711649	0.124611139297485\\
-0.0142314834520221	0.098982147872448\\
};
\addplot [color=mycolor1,solid,forget plot]
  table[row sep=crcr]{%
-0.654860734939575	0.755749583244324\\
-0.520174384117126	0.600313246250153\\
-0.413189202547073	0.476845741271973\\
-0.328207850456238	0.378772050142288\\
-0.260704755783081	0.300869315862656\\
-0.207085147500038	0.238988995552063\\
-0.164493575692177	0.189835712313652\\
-0.130661889910698	0.150791868567467\\
-0.103788435459137	0.119778238236904\\
-0.082442082464695	0.095143236219883\\
-0.0654860734939575	0.0755749568343163\\
};
\addplot [color=mycolor1,solid,forget plot]
  table[row sep=crcr]{%
-0.959492981433868	0.281732559204102\\
-0.762152373790741	0.223788127303123\\
-0.605399131774902	0.177761226892471\\
-0.480885624885559	0.141200765967369\\
-0.381981045007706	0.112159751355648\\
-0.303418308496475	0.0890916585922241\\
-0.241013735532761	0.0707680210471153\\
-0.191444024443626	0.0562130361795425\\
-0.152069389820099	0.0446516014635563\\
-0.120793007314205	0.0354680269956589\\
-0.095949299633503	0.0281732548028231\\
};
\addplot [color=mycolor1,solid,forget plot]
  table[row sep=crcr]{%
-0.959492981433868	-0.281732559204102\\
-0.762152373790741	-0.223788127303123\\
-0.605399131774902	-0.177761226892471\\
-0.480885624885559	-0.141200765967369\\
-0.381981045007706	-0.112159751355648\\
-0.303418308496475	-0.0890916585922241\\
-0.241013735532761	-0.0707680210471153\\
-0.191444024443626	-0.0562130361795425\\
-0.152069389820099	-0.0446516014635563\\
-0.120793007314205	-0.0354680269956589\\
-0.095949299633503	-0.0281732548028231\\
};
\addplot [color=mycolor1,solid,forget plot]
  table[row sep=crcr]{%
-0.654860734939575	-0.755749583244324\\
-0.520174384117126	-0.600313246250153\\
-0.413189202547073	-0.476845741271973\\
-0.328207850456238	-0.378772050142288\\
-0.260704755783081	-0.300869315862656\\
-0.207085147500038	-0.238988995552063\\
-0.164493575692177	-0.189835712313652\\
-0.130661889910698	-0.150791868567467\\
-0.103788435459137	-0.119778238236904\\
-0.082442082464695	-0.095143236219883\\
-0.0654860734939575	-0.0755749568343163\\
};
\addplot [color=mycolor1,solid,forget plot]
  table[row sep=crcr]{%
-0.142314836382866	-0.989821434020996\\
-0.113044694066048	-0.786243140697479\\
-0.0897945910692215	-0.624535083770752\\
-0.07132638245821	-0.49608588218689\\
-0.0566565580666065	-0.394055008888245\\
-0.0450039021670818	-0.313009023666382\\
-0.0357478708028793	-0.248631909489632\\
-0.0283955428749323	-0.197495341300964\\
-0.0225553810596466	-0.156876131892204\\
-0.0179163757711649	-0.124611139297485\\
-0.0142314834520221	-0.098982147872448\\
};
\addplot [color=mycolor1,solid,forget plot]
  table[row sep=crcr]{%
0.415415018796921	-0.909631967544556\\
0.329975873231888	-0.722546398639679\\
0.262109160423279	-0.573938965797424\\
0.208200708031654	-0.455895930528641\\
0.16537968814373	-0.362131029367447\\
0.131365761160851	-0.287650883197784\\
0.104347534477711	-0.228489220142365\\
0.0828861892223358	-0.181495442986488\\
0.0658388435840607	-0.144166961312294\\
0.0522976517677307	-0.114515885710716\\
0.0415415018796921	-0.0909631997346878\\
};
\addplot [color=mycolor1,solid,forget plot]
  table[row sep=crcr]{%
0.841253519058228	-0.540640830993652\\
0.668231427669525	-0.429446280002594\\
0.530795097351074	-0.341121286153793\\
0.421625524759293	-0.27096226811409\\
0.334909051656723	-0.215232983231544\\
0.266027718782425	-0.170965641736984\\
0.211313337087631	-0.135802835226059\\
0.167852148413658	-0.107872024178505\\
0.133329704403877	-0.0856857970356941\\
0.105907544493675	-0.0680626481771469\\
0.084125354886055	-0.054064080119133\\
};
\addplot [color=mycolor1,solid,forget plot]
  table[row sep=crcr]{%
1	-1.11022302462516e-15\\
0.794328212738037	-8.88178419700125e-16\\
0.630957365036011	-6.66133814775094e-16\\
0.501187205314636	-5.55111512312578e-16\\
0.398107171058655	-4.44089209850063e-16\\
0.31622776389122	-3.33066907387547e-16\\
0.251188635826111	-3.33066907387547e-16\\
0.199526235461235	-2.22044604925031e-16\\
0.158489316701889	-2.22044604925031e-16\\
0.125892534852028	-1.11022302462516e-16\\
0.100000001490116	-1.11022302462516e-16\\
};
\addplot [color=red,mark size=3.3pt,only marks,mark=*,mark options={solid},forget plot]
  table[row sep=crcr]{%
1	0\\
};

\addplot[area legend,solid,line width=1.5pt,draw=black,fill=black,forget plot]
table[row sep=crcr] {%
x	y\\
0.928329110145569	0.244088590145111\\
1	0\\
1	0\\
1	0\\
1	0\\
1	0\\
0.856658220291138	0.488177180290222\\
0.837310075759888	0.478587985038757\\
0.841253519058228	0.540640830993652\\
0.878119349479675	0.490570783615112\\
0.856658220291138	0.488177180290222\\
}--cycle;

\addplot[area legend,solid,line width=1.5pt,draw=black,fill=black,forget plot]
table[row sep=crcr] {%
x	y\\
0.924503326416016	0\\
1	0\\
1	0\\
1	0\\
1	0\\
1	0\\
0.849006652832031	0\\
0.852756500244141	-0.0212661027908325\\
0.794328212738037	0\\
0.85275661945343	0.0212662220001221\\
0.849006652832031	0\\
}--cycle;
\end{axis}
\end{tikzpicture}%

%% file: figures/warp-sphere.tikz
%
%
\definecolor{mycolor1}{rgb}{0.00000,0.44700,0.74100}%
\begin{tikzpicture}

\begin{axis}[%
width=3.744in,
height=3.744in,
at={(1.427in,0.505in)},
scale only axis,
xmin=-0.884468078613281,
xmax=0.884468078613281,
ymin=-0.884468078613281,
ymax=0.884468078613281,
hide axis,
axis x line*=bottom,
axis y line*=left
]
\addplot [color=mycolor1,solid,forget plot]
  table[row sep=crcr]{%
-0.247392252087593	-0.80057817697525\\
-0.439636498689651	-0.728407025337219\\
-0.611042857170105	-0.603336870670319\\
-0.746869444847107	-0.431725084781647\\
-0.834282100200653	-0.225230813026428\\
-0.864468097686768	-1.11022302462516e-16\\
-0.834282100200653	0.225230813026428\\
-0.746869444847107	0.431725084781647\\
-0.611042857170105	0.603336870670319\\
-0.439636498689651	0.728407025337219\\
-0.247392252087593	0.80057817697525\\
};
\addplot [color=mycolor1,solid,forget plot]
  table[row sep=crcr]{%
-0.225090146064758	-0.820482492446899\\
-0.407884210348129	-0.761223971843719\\
-0.577046990394592	-0.641792297363281\\
-0.715453326702118	-0.465842485427856\\
-0.806649804115295	-0.245298609137535\\
-0.838539302349091	-1.11022302462516e-16\\
-0.806649804115295	0.245298609137535\\
-0.715453326702118	0.465842485427856\\
-0.577046990394592	0.641792297363281\\
-0.407884210348129	0.761223971843719\\
-0.225090146064758	0.820482492446899\\
};
\addplot [color=mycolor1,solid,forget plot]
  table[row sep=crcr]{%
-0.186441347002983	-0.838229298591614\\
-0.343889534473419	-0.791593730449677\\
-0.494696944952011	-0.678625822067261\\
-0.62191241979599	-0.499453186988831\\
-0.707693994045258	-0.265438169240952\\
-0.738067030906677	-1.11022302462516e-16\\
-0.707693994045258	0.265438169240952\\
-0.62191241979599	0.499453186988831\\
-0.494696944952011	0.678625822067261\\
-0.343889534473419	0.791593730449677\\
-0.186441347002983	0.838229298591614\\
};
\addplot [color=mycolor1,solid,forget plot]
  table[row sep=crcr]{%
-0.133410394191742	-0.852292358875275\\
-0.249610885977745	-0.816441774368286\\
-0.364085704088211	-0.709697186946869\\
-0.463166803121567	-0.528544187545776\\
-0.531320929527283	-0.283173799514771\\
-0.555717945098877	-1.11022302462516e-16\\
-0.531320929527283	0.283173799514771\\
-0.463166803121567	0.528544187545776\\
-0.364085704088211	0.709697186946869\\
-0.249610885977745	0.816441774368286\\
-0.133410394191742	0.852292358875275\\
};
\addplot [color=mycolor1,solid,forget plot]
  table[row sep=crcr]{%
-0.0696001499891281	-0.86134272813797\\
-0.131437569856644	-0.83281421661377\\
-0.193496108055115	-0.730648696422577\\
-0.24814710021019	-0.54855477809906\\
-0.286256223917007	-0.295541167259216\\
-0.300002038478851	-1.11022302462516e-16\\
-0.286256223917007	0.295541167259216\\
-0.24814710021019	0.54855477809906\\
-0.193496108055115	0.730648696422577\\
-0.131437569856644	0.83281421661377\\
-0.0696001499891281	0.86134272813797\\
};
\addplot [color=mycolor1,solid,forget plot]
  table[row sep=crcr]{%
0	-0.864468097686768\\
0	-0.838539302349091\\
0	-0.738067030906677\\
0	-0.555717945098877\\
0	-0.300002038478851\\
0	-1.11022302462516e-16\\
0	0.300002038478851\\
0	0.555717945098877\\
0	0.738067030906677\\
0	0.838539302349091\\
0	0.864468097686768\\
};
\addplot [color=mycolor1,solid,forget plot]
  table[row sep=crcr]{%
0.0696001499891281	-0.86134272813797\\
0.131437569856644	-0.83281421661377\\
0.193496108055115	-0.730648696422577\\
0.24814710021019	-0.54855477809906\\
0.286256223917007	-0.295541167259216\\
0.300002038478851	-1.11022302462516e-16\\
0.286256223917007	0.295541167259216\\
0.24814710021019	0.54855477809906\\
0.193496108055115	0.730648696422577\\
0.131437569856644	0.83281421661377\\
0.0696001499891281	0.86134272813797\\
};
\addplot [color=mycolor1,solid,forget plot]
  table[row sep=crcr]{%
0.133410394191742	-0.852292358875275\\
0.249610885977745	-0.816441774368286\\
0.364085704088211	-0.709697186946869\\
0.463166803121567	-0.528544187545776\\
0.531320929527283	-0.283173799514771\\
0.555717945098877	-1.11022302462516e-16\\
0.531320929527283	0.283173799514771\\
0.463166803121567	0.528544187545776\\
0.364085704088211	0.709697186946869\\
0.249610885977745	0.816441774368286\\
0.133410394191742	0.852292358875275\\
};
\addplot [color=mycolor1,solid,forget plot]
  table[row sep=crcr]{%
0.186441347002983	-0.838229298591614\\
0.343889534473419	-0.791593730449677\\
0.494696944952011	-0.678625822067261\\
0.62191241979599	-0.499453186988831\\
0.707693994045258	-0.265438169240952\\
0.738067030906677	-1.11022302462516e-16\\
0.707693994045258	0.265438169240952\\
0.62191241979599	0.499453186988831\\
0.494696944952011	0.678625822067261\\
0.343889534473419	0.791593730449677\\
0.186441347002983	0.838229298591614\\
};
\addplot [color=mycolor1,solid,forget plot]
  table[row sep=crcr]{%
0.225090146064758	-0.820482492446899\\
0.407884210348129	-0.761223971843719\\
0.577046990394592	-0.641792297363281\\
0.715453326702118	-0.465842485427856\\
0.806649804115295	-0.245298609137535\\
0.838539302349091	-1.11022302462516e-16\\
0.806649804115295	0.245298609137535\\
0.715453326702118	0.465842485427856\\
0.577046990394592	0.641792297363281\\
0.407884210348129	0.761223971843719\\
0.225090146064758	0.820482492446899\\
};
\addplot [color=mycolor1,solid,forget plot]
  table[row sep=crcr]{%
0.247392252087593	-0.80057817697525\\
0.439636498689651	-0.728407025337219\\
0.611042857170105	-0.603336870670319\\
0.746869444847107	-0.431725084781647\\
0.834282100200653	-0.225230813026428\\
0.864468097686768	-1.11022302462516e-16\\
0.834282100200653	0.225230813026428\\
0.746869444847107	0.431725084781647\\
0.611042857170105	0.603336870670319\\
0.439636498689651	0.728407025337219\\
0.247392252087593	0.80057817697525\\
};
\addplot [color=mycolor1,solid,forget plot]
  table[row sep=crcr]{%
-0.247392252087593	-0.80057817697525\\
-0.225090146064758	-0.820482492446899\\
-0.186441347002983	-0.838229298591614\\
-0.133410394191742	-0.852292358875275\\
-0.0696001499891281	-0.86134272813797\\
0	-0.864468097686768\\
0.0696001499891281	-0.86134272813797\\
0.133410394191742	-0.852292358875275\\
0.186441347002983	-0.838229298591614\\
0.225090146064758	-0.820482492446899\\
0.247392252087593	-0.80057817697525\\
};
\addplot [color=mycolor1,solid,forget plot]
  table[row sep=crcr]{%
-0.439636498689651	-0.728407025337219\\
-0.407884210348129	-0.761223971843719\\
-0.343889534473419	-0.791593730449677\\
-0.249610885977745	-0.816441774368286\\
-0.131437569856644	-0.83281421661377\\
0	-0.838539302349091\\
0.131437569856644	-0.83281421661377\\
0.249610885977745	-0.816441774368286\\
0.343889534473419	-0.791593730449677\\
0.407884210348129	-0.761223971843719\\
0.439636498689651	-0.728407025337219\\
};
\addplot [color=mycolor1,solid,forget plot]
  table[row sep=crcr]{%
-0.611042857170105	-0.603336870670319\\
-0.577046990394592	-0.641792297363281\\
-0.494696944952011	-0.678625822067261\\
-0.364085704088211	-0.709697186946869\\
-0.193496108055115	-0.730648696422577\\
0	-0.738067030906677\\
0.193496108055115	-0.730648696422577\\
0.364085704088211	-0.709697186946869\\
0.494696944952011	-0.678625822067261\\
0.577046990394592	-0.641792297363281\\
0.611042857170105	-0.603336870670319\\
};
\addplot [color=mycolor1,solid,forget plot]
  table[row sep=crcr]{%
-0.746869444847107	-0.431725084781647\\
-0.715453326702118	-0.465842485427856\\
-0.62191241979599	-0.499453186988831\\
-0.463166803121567	-0.528544187545776\\
-0.24814710021019	-0.54855477809906\\
0	-0.555717945098877\\
0.24814710021019	-0.54855477809906\\
0.463166803121567	-0.528544187545776\\
0.62191241979599	-0.499453186988831\\
0.715453326702118	-0.465842485427856\\
0.746869444847107	-0.431725084781647\\
};
\addplot [color=mycolor1,solid,forget plot]
  table[row sep=crcr]{%
-0.834282100200653	-0.225230813026428\\
-0.806649804115295	-0.245298609137535\\
-0.707693994045258	-0.265438169240952\\
-0.531320929527283	-0.283173799514771\\
-0.286256223917007	-0.295541167259216\\
0	-0.300002038478851\\
0.286256223917007	-0.295541167259216\\
0.531320929527283	-0.283173799514771\\
0.707693994045258	-0.265438169240952\\
0.806649804115295	-0.245298609137535\\
0.834282100200653	-0.225230813026428\\
};
\addplot [color=mycolor1,solid,forget plot]
  table[row sep=crcr]{%
-0.864468097686768	-1.11022302462516e-16\\
-0.838539302349091	-1.11022302462516e-16\\
-0.738067030906677	-1.11022302462516e-16\\
-0.555717945098877	-1.11022302462516e-16\\
-0.300002038478851	-1.11022302462516e-16\\
0	-1.11022302462516e-16\\
0.300002038478851	-1.11022302462516e-16\\
0.555717945098877	-1.11022302462516e-16\\
0.738067030906677	-1.11022302462516e-16\\
0.838539302349091	-1.11022302462516e-16\\
0.864468097686768	-1.11022302462516e-16\\
};
\addplot [color=mycolor1,solid,forget plot]
  table[row sep=crcr]{%
-0.834282100200653	0.225230813026428\\
-0.806649804115295	0.245298609137535\\
-0.707693994045258	0.265438169240952\\
-0.531320929527283	0.283173799514771\\
-0.286256223917007	0.295541167259216\\
0	0.300002038478851\\
0.286256223917007	0.295541167259216\\
0.531320929527283	0.283173799514771\\
0.707693994045258	0.265438169240952\\
0.806649804115295	0.245298609137535\\
0.834282100200653	0.225230813026428\\
};
\addplot [color=mycolor1,solid,forget plot]
  table[row sep=crcr]{%
-0.746869444847107	0.431725084781647\\
-0.715453326702118	0.465842485427856\\
-0.62191241979599	0.499453186988831\\
-0.463166803121567	0.528544187545776\\
-0.24814710021019	0.54855477809906\\
0	0.555717945098877\\
0.24814710021019	0.54855477809906\\
0.463166803121567	0.528544187545776\\
0.62191241979599	0.499453186988831\\
0.715453326702118	0.465842485427856\\
0.746869444847107	0.431725084781647\\
};
\addplot [color=mycolor1,solid,forget plot]
  table[row sep=crcr]{%
-0.611042857170105	0.603336870670319\\
-0.577046990394592	0.641792297363281\\
-0.494696944952011	0.678625822067261\\
-0.364085704088211	0.709697186946869\\
-0.193496108055115	0.730648696422577\\
0	0.738067030906677\\
0.193496108055115	0.730648696422577\\
0.364085704088211	0.709697186946869\\
0.494696944952011	0.678625822067261\\
0.577046990394592	0.641792297363281\\
0.611042857170105	0.603336870670319\\
};
\addplot [color=mycolor1,solid,forget plot]
  table[row sep=crcr]{%
-0.439636498689651	0.728407025337219\\
-0.407884210348129	0.761223971843719\\
-0.343889534473419	0.791593730449677\\
-0.249610885977745	0.816441774368286\\
-0.131437569856644	0.83281421661377\\
0	0.838539302349091\\
0.131437569856644	0.83281421661377\\
0.249610885977745	0.816441774368286\\
0.343889534473419	0.791593730449677\\
0.407884210348129	0.761223971843719\\
0.439636498689651	0.728407025337219\\
};
\addplot [color=mycolor1,solid,forget plot]
  table[row sep=crcr]{%
-0.247392252087593	0.80057817697525\\
-0.225090146064758	0.820482492446899\\
-0.186441347002983	0.838229298591614\\
-0.133410394191742	0.852292358875275\\
-0.0696001499891281	0.86134272813797\\
0	0.864468097686768\\
0.0696001499891281	0.86134272813797\\
0.133410394191742	0.852292358875275\\
0.186441347002983	0.838229298591614\\
0.225090146064758	0.820482492446899\\
0.247392252087593	0.80057817697525\\
};
\addplot [color=red,mark size=3.3pt,only marks,mark=*,mark options={solid},forget plot]
  table[row sep=crcr]{%
0	-1.11022302462516e-16\\
};

\addplot[area legend,solid,line width=1.5pt,draw=black,fill=black,forget plot]
table[row sep=crcr] {%
x	y\\
0	0.12396365404129\\
0	0\\
0	0\\
0	0\\
0	0\\
0	0\\
0	0.247927308082581\\
-0.0202535390853882	0.244356155395508\\
0	0.300002098083496\\
0.0202535390853882	0.244356155395508\\
0	0.247927308082581\\
}--cycle;

\addplot[area legend,solid,line width=1.5pt,draw=black,fill=black,forget plot]
table[row sep=crcr] {%
x	y\\
0.12396365404129	0\\
0	0\\
0	0\\
0	0\\
0	0\\
0	0\\
0.247927308082581	0\\
0.244356155395508	-0.0202535390853882\\
0.300002098083496	0\\
0.244356155395508	0.0202535390853882\\
0.247927308082581	0\\
}--cycle;
\end{axis}
\end{tikzpicture}%